\RequirePackage{fix-cm}

\documentclass[twocolumn]{svjour3}          
\smartqed  
\usepackage{graphicx}
\usepackage{multirow}
\graphicspath{{figure/}}
\usepackage{subfig}
\usepackage{microtype}
\usepackage{natbib}
\usepackage{algorithm,algorithmic}
\usepackage{amsmath,amssymb} 
\usepackage{pifont}
\usepackage{bm}

\usepackage[american]{babel}
\usepackage{url}

\begin{document}

\title{Unsupervised Object Discovery and Co-Localization\\by Deep Descriptor Transforming}


\author{Xiu-Shen Wei \and Chen-Lin Zhang \and Jianxin Wu \and Chunhua Shen \and Zhi-Hua Zhou\thanks{The first two authors contributed equally to this work. This work was done when X.-S. Wei was visiting the University of Adelaide.}\thanks{Corresponding authors: Jianxin Wu and Chunhua Shen}
}


\institute{Xiu-Shen Wei \and Chen-Lin Zhang \and Jianxin Wu \and Zhi-Hua Zhou\at
              Nanjing University, China \\
              \email{\{weixs, zhangcl, wujx, zhouzh\}@lamda.nju.edu.cn}
           \and
              Chunhua Shen \at
              The University of Adelaide, Australia \\
              \email{chunhua.shen@adelaide.edu.au}
}

\date{Received: date / Accepted: date}

\maketitle

\begin{abstract}
Reusable model design becomes desirable with the rapid expansion of computer vision and machine learning applications. In this paper, we focus on the reusability of pre-trained deep convolutional models. Specifically, different from treating pre-trained models as feature extractors, we reveal more treasures beneath convolutional layers, i.e., the convolutional activations could act as a detector for the common object in the image co-localization problem. We propose a simple yet effective method, termed Deep Descriptor Transforming (DDT), for evaluating the correlations of descriptors and then obtaining the category-consistent regions, which can accurately locate the common object in a set of unlabeled images, i.e., unsupervised object discovery. Empirical studies validate the effectiveness of the proposed DDT method. On benchmark image co-localization datasets, DDT consistently outperforms existing state-of-the-art methods by a large margin. Moreover, DDT also demonstrates good generalization ability for unseen categories and robustness for dealing with noisy data. Beyond those, DDT can be also employed for harvesting web images into valid external data sources for improving performance of both image recognition and object detection.
\end{abstract}

\section{Introduction}

Model reuse~\citep{learnware} attempts to construct a model by utilizing existing available models, mostly trained for other tasks, rather than building a model from scratch. Particularly in deep learning, since deep convolutional neural networks have achieved great success in various tasks involving images, videos, texts and more, there are several studies have the flavor of reusing deep models pre-trained on \emph{ImageNet}~\citep{russaijcv2015}.

In computer vision, pre-trained models on \emph{ImageNet} have been successfully adopted to various usages, e.g., as universal feature extractors~\citep{majiijcv2016,wangiccv2015, yaoeccv2016}, object proposal generators~\citep{Amir15ICCV}, etc. In particular, \citet{scda2016} proposed the SCDA (Selective Convolutional Descriptor Aggregation) method to utilize pre-trained models for both localizing a single fine-grained object (e.g., birds of different species) in each image and retrieving fine-grained images of the same classes/species in an unsupervised fashion. 

\begin{figure}[t]
 \centering
 \includegraphics[width=0.95\columnwidth]{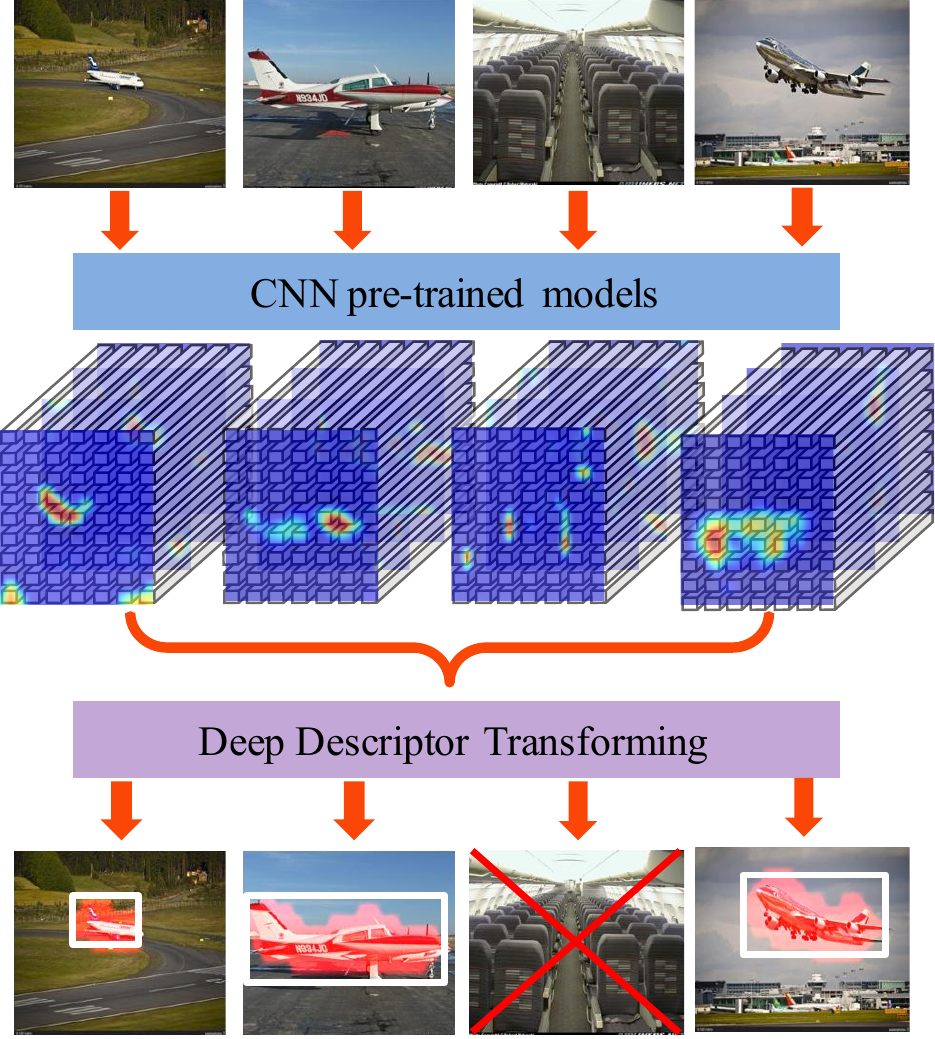}
 \caption{Pipeline of the proposed DDT method for image co-localization. In this instance, the goal is to localize the \emph{airplane} within each image. Note that, there might be few noisy images in the image set. (Best viewed in color.)}
 \label{fig:pipeline}
\end{figure}

In this paper, we reveal that the convolutional activations can be used as a detector for the \emph{common object} in image co-localization. Image co-localization (a.k.a. unsupervised object discovery) is a fundamental computer vision problem, which simultaneously localizes objects of the same category across a set of distinct images. Specifically, we propose a simple but effective method termed Deep Descriptor Transforming (DDT) for image co-localization. In DDT, the deep convolutional descriptors extracted from pre-trained deep convolutional models are transformed into a new space, where it can evaluate the correlations between these descriptors. By leveraging the correlations among images in the image set, the common object inside these images can be located automatically without additional supervision signals. The pipeline of DDT is shown in Fig.~\ref{fig:pipeline}. To our knowledge, this is \emph{the first work} to demonstrate the possibility of convolutional activations/descriptors in pre-trained models \emph{being able to act as a detector for the common object.}

Experimental results show that DDT significantly outperforms existing state-of-the-art methods, including image co-localization and weakly supervised object localization, in both the deep learning and hand-crafted feature scenarios. Besides, we empirically show that DDT has a good generalization ability for unseen images apart from \emph{ImageNet}. More importantly, the proposed method is robust, because DDT can also detect the noisy images which do not contain the common object. Thanks to the advantages of DDT, our method could be used as a tool to harvest easy-to-obtain but noisy web images. We can employ DDT to remove noisy images from webly datasets for improving image recognition accuracy. Moreover, it can be also utilized to supply object bounding boxes of web images. Then, we use these images with automatically labeled object boxes as valid external data sources to enhance object detection performance.

Our main contributions are as follows:
\begin{enumerate}
\item[1.] We propose a simple yet effective method, i.e., Deep Descriptor Transforming, for unsupervised object discovery and co-localization. Besides, DDT reveals another probability of deep pre-trained network reusing, i.e., convolutional activations/descriptors can play a role as a common object detector. 
\item[2.] The co-localization process of DDT is both effective and efficient, which does not require image labels, negative images or redundant object proposals. DDT consistently outperforms state-of-the-arts of image co-localization methods and weakly supervised object localization methods. With the ensemble of multiple CNN layers, DDT could further improve its co-localization performance.
\item[3.] DDT has a good generalization ability for unseen categories and robustness for dealing with noisy data. Thanks to these advantages, DDT can be employed beyond the narrow co-localization task. Specifically, it can be used as a generalized tool for exploiting noisy but free web images. By removing noisy images and automatically supplying object bounding boxes, these web images processed by DDT could become valid external data sources for improving both recognition and detection performance. We thus provide a very useful tool for automatically annotating images. The effectiveness of DDT augmentation on recognition and detection is validated in Sec.~\ref{sec:web}.
\item[4.] Based on the previous point, we also collect an object detection dataset from web images, named \emph{WebVOC}. It shares the same 20 categories as the \emph{PASCAL VOC} dataset \citep{voc2015}, and has a similar dataset scale (10k images) comparing with \emph{PASCAL VOC}. We also release the \emph{WebVOC} dataset with the automatically generated bounding boxes by DDT for further study. 
\end{enumerate}

This paper is extended based on our preliminary work \citep{ddtijcai}. Comparing with it, we now further introduce the multiple layer ensemble strategy for improving co-localization performance, provide DDT augmentation for handling web images, apply the proposed method on webly-supervised learning tasks (i.e., both recognition and detection), and supply our DDT based webly object detection dataset.

The remainder of the paper is organized as follows. In Sec.~\ref{sec:related}, we briefly review related literature of CNN model reuse, image co-localization and webly-supervised learning. In Sec.~\ref{sec:ourddt}, we introduce our proposed method (DDT and its variant DDT$^+$). Sec.~\ref{sec:experi} reports the image co-localization results and the results of webly-supervised learning tasks. We conclude the paper in Sec.~\ref{sec:conc} finally.

\section{Related Work}\label{sec:related}

We briefly review three lines of related work: model reuse of CNNs, research on image co-localization and webly-supervised learning.

\subsection{CNN Model Reuse}

Reusability has been emphasized by~\citep{learnware} as a crucial characteristic of the new concept of \emph{learnware}. It would be ideal if models can be reused in scenarios that are very different from their original training scenarios. Particularly, with the breakthrough in image classification using Convolutional Neural Networks (CNN), pre-trained CNN models trained for one task (e.g., recognition) have also been applied to domains different from their original purposes (e.g., for describing texture \citep{majiijcv2016} or finding object proposals~\citep{Amir15ICCV}). However, for such adaptations of pre-trained models, they still require further annotations in the new domain (e.g., image labels). While, DDT deals with the image co-localization problem in an unsupervised setting.

Coincidentally, several recent works also shed lights on CNN pre-trained model reuse in the unsupervised setting, e.g., SCDA (Selective Convolutional Descriptor Aggregation)~\citep{scda2016}. SCDA is proposed for handling the fine-grained image retrieval task, where it uses pre-trained models (from \emph{ImageNet}) to locate main objects in fine-grained images. It is the most related work to ours, even though SCDA is not for image co-localization. Different from our DDT, SCDA assumes only an object of interest in each image, and meanwhile objects from other categories does not exist. Thus, SCDA locates the object using cues from this \emph{single} image assumption. Clearly, it can not work well for images containing diverse objects (cf. Table~\ref{table:voc07} and Table~\ref{table:voc12}), and also can not handle data noise (cf. Sec.~\ref{sec:noise}).

\subsection{Image Co-Localization}

Image co-localization, a.k.a. unsupervised object discovery~\citep{chicvpr2015,wangiccv2015}, is a fundamental problem in computer vision, where it needs to discover the common object emerging in only positive sets of example images (without any negative examples or further supervisions). Image co-localization shares some similarities with image co-segmentation~\citep{cosegijcai,gunheeiccv2011,joulinmccvpr2010}. Instead of generating a precise segmentation of the related objects in each image, co-localization methods aim to return a bounding box around the object. Moreover, it also allows us to extract rich features from within the boxes to compare across images, which has shown to be very helpful for detection~\citep{tangcvpr2014}.

Additionally, co-localization is also related to weakly supervised object localization (WSOL)~\citep{wsolijcai,bilencvpr2015,wangeccv2014,sivacvpr2011}. But the key difference between them is that WSOL requires manually-labeled negative images whereas co-localization does not. Thus, WSOL methods could achieve better localization performance than co-localization methods. However, our proposed methods perform comparably with state-of-the-art WSOL methods and even outperform them (cf. Table~\ref{table:voc07weak}).

In the literature, some representative co-localization methods are based on low-level visual cues and optimization algorithms. \citet{tangcvpr2014} formulated co-localization as a boolean constrained quadratic program which can be relaxed to a convex problem. Then, it was further accelerated by the Frank-Wolfe algorithm~\citep{joulineccv2014}. After that, \citet{chicvpr2015} proposed a Probabilistic Hough Matching algorithm to match object proposals across images and then dominant objects are localized by selecting proposals based on matching scores.

Recently, there also emerge several co-localization methods based on pre-trained deep convolutional models, e.g., \citet{yaoeccv2016,wangeccv2014}. Unfortunately, these methods just treated pre-trained models as simple feature extractors to extract the fully connected representations, which did not sufficiently mine the treasures beneath the convolutional layers (i.e., leveraging the original correlations between deep descriptors among convolutional layers). Moreover, these methods also require object proposals as a part of their object discovery, which not only made them highly depend on the quality of object proposals, but may lead to huge computational costs. In addition, almost all the previous co-localization methods can not handle noisy data, except for~\citep{tangcvpr2014}.

Comparing with previous works, our DDT is unsupervised, without utilizing bounding boxes, additional image labels or redundant object proposals. Images only need one forward run through a pre-trained model. Then, efficient deep descriptor transforming is employed for obtaining the category-consistent image regions. DDT is very easy to implement, and surprisingly has good generalization ability and robustness. Furthermore, DDT can be used a valid data augmentation tool for handling noisy but free web images.

\subsection{Webly-Supervised Learning}

Recent development of deep CNNs has led to great success in a variety of computer vision tasks. This success is largely driven by the availability of large scale well-annotated image datasets, e.g., \emph{ImageNet}~\citep{russaijcv2015}, \emph{MS COCO}~\citep{coco} and \emph{PASCAL VOC}~\citep{voc2015}. However, annotating a massive number of images is extremely labor-intensive and costly. To reduce the annotation labor costs, an alternative approach is to obtain the image annotations directly from the image search engine from the Internet, e.g., Google or Bing.

However, the annotations of web images returned by a search engine will inevitably be noisy since the query keywords may not be consistent with the visual content of target images. Thus, webly-supervised learning methods are proposed for overcoming this issue.

There are two main branches of webly-supervised learning. The first branch attempts to boost existing object recognition task performance using web resources \citep{bohan2017,Georgeiccv2015,rubinscvpr2015}. Some work was implemented as semi-supervised frameworks by first generating a small group of labeled seed images and then enlarging the dataset from these seeds via web data, e.g., \citet{Georgeiccv2015,rubinscvpr2015}. In very recently, \citet{bohan2017} proposed a two-level attention framework for dealing with webly-supervised classification, which achieves a new state-of-the-art. Specifically, they not only used a high-level attention focusing on a group of images for filtering out noisy images, but also employed a low-level attention for capturing the discriminative image regions on the single image level

The second branch is learning visual concepts directly from the web, e.g., \citet{Ferguseccv2004,xinjinpami2016}. Methods belonging to this category usually collected a large image pool from image search engines and then performed a filtering operation to remove noise and discover visual concepts. Our strategy for handling web data based on DDT naturally falls into the second category. In practice, since DDT could (1) recognize noisy images and also (2) supply bounding boxes of objects, we leverage the first usage of DDT to handle webly-supervised classification (cf. Table~\ref{table:webcar} and Table~\ref{table:webimagenet}), and leverage both two usages to deal with webly-supervised detection (cf. Table~\ref{table:webdetect07} and Table~\ref{table:webdetect12}).

\section{The Proposed Method}\label{sec:ourddt}

In this section, we propose the Deep Descriptor Transforming (DDT) method. Firstly, we introduce notations used in this paper. Then, we present the DDT process followed by discussions and analyses. Finally, in order to further improve the image co-localization performance, the multiple layer ensemble strategy is utilized in DDT.

\subsection{Preliminaries}

The following notation is used in the rest of this paper. The term ``feature map'' indicates the convolution results of one channel; the term ``activations'' indicates feature maps of all channels in a convolution layer; and the term ``descriptor'' indicates the $d$-dimensional component vector of activations.

Given an input image $I$ of size $H\times W$, the activations of a convolution layer are formulated as an order-3 tensor $T$ with $h\times w\times d$ elements. $T$ can be considered as having $h\times w$ cells and each cell contains one $d$-dimensional deep descriptor. For the $n$-th image in the image set, we denote its corresponding deep descriptors as $X^n=\left \{\bm{x}^n_{\left(i,j\right)}\in \mathcal{R}^{d}\right \}$, where $\left(i,j \right)$ is a particular cell ($i\in \left \{1,\ldots,h\right \}, j\in \left \{1,\ldots,w \right \}$) and $n\in \left \{1,\ldots,N\right \}$.

\subsection{SCDA Recap}\label{sec:scda}

Since SCDA (Selective Convolutional Descriptor Aggregation) \citep{scda2016} is the most related work to ours, we hereby present a recap of this method. SCDA is proposed for dealing with the fine-grained image retrieval problem. It employs pre-trained CNN models to select the meaningful deep descriptors by localizing the main object in fine-grained images unsupervisedly. In SCDA, it assumes that each image contains only one main object of interest and without other categories' objects. Thus, the object localization strategy is based on the activation tensor of a \emph{single} image.

Concretely, for an image, the activation tensor is added up through the depth direction. Thus, the $h\times w\times d$ 3-D tensor becomes a $h\times w$ 2-D matrix, which is called the ``aggregation map'' in SCDA. Then, the mean value $\bar{a}$ of the aggregation map is regarded as the threshold for localizing the object. If the activation response in the position $\left(i,j\right)$ of the aggregation map is larger than $\bar{a}$, it indicates the object might appear in that position.

\subsection{Deep Descriptor Transforming (DDT)}

What distinguishes DDT from SCDA is that we can leverage the correlations beneath the whole \emph{image set}, instead of a \emph{single} image. Additionally, different from weakly supervised object localization, we do not have either image labels or negative image sets in WSOL, so that the information we can use is only from the pre-trained models. Here, we transform the deep descriptors in convolutional layers to mine the hidden cues for co-localizing common objects.

Principal component analysis (PCA)~\citep{pca1901} is a statistical procedure, which uses an orthogonal transformation to convert a set of observations of possibly correlated variables into a set of linearly uncorrelated variables (i.e., the principal components). This transformation is defined in such a way that the first principal component has the largest possible variance, and each succeeding component in turn has the highest variance possible under the constraint that it is orthogonal to all the preceding components.

PCA is widely used in computer vision and machine learning for image denoising~\citep{stereijcv2017}, 3D object retrieval~\citep{3dijcv2011}, statistical shape modeling~\citep{shellpcaiccv2015}, subspace learning~\citep{videosubsijcv2013,subsijcv2003}, and so on. Specifically, in this paper, we utilize PCA as projection directions for transforming these deep descriptors $\{\bm{x}^{\cdot}_{\left(i,j\right)}\}$ to evaluate their correlations. Then, on each projection direction, the corresponding principal component's values are treated as the cues for image co-localization, especially the first principal component. Thanks to the property of this kind of transforming, DDT is also able to handle data noise.

In DDT, for a set of $N$ images containing objects from the same category, we first collect the corresponding convolutional descriptors ($X^1,\ldots,X^N$) from the last convolutional layer by feeding the images into a pre-trained CNN model. Then, the mean vector of all the descriptors is calculated by:
\begin{equation}
\label{eq:mean}
\bar{\bm{x}} = \frac{1}{K} \sum_n \sum_{i,j} \bm{x}_{\left(i,j\right)}^n \,,
\end{equation}
where $K=h\times w\times N$. Note that, here we assume each image has the same number of deep descriptors (i.e., $h\times w$) for presentation clarity. Our proposed method, however, can handle input images with arbitrary resolutions.

Then, after obtaining the covariance matrix:
\begin{equation}
\label{eq:cov}
{\rm{Cov}} (\bm{x})=\frac{1}{K} \sum_n \sum_{i,j} (\bm{x}_{\left(i,j\right)}^n-\bar{\bm{x}}) (\bm{x}_{\left(i,j\right)}^n-\bar{\bm{x}})^\top \,,
\end{equation}
we can get the eigenvectors $\bm{\xi}_1,\ldots, \bm{\xi}_d$ of ${\rm Cov}(\bm{x})$ which correspond to the sorted eigenvalues $\lambda_1\geq \cdots \geq \lambda_d \geq 0$.

As aforementioned, since the first principal component has the largest variance, we take the eigenvector $\bm{\xi}_1$ corresponding to the largest eigenvalue as the main projection direction. For the deep descriptor at a particular position $\left(i,j\right)$ of an image, its first principal component $p^1$ is calculated as follows:
\begin{equation}
\label{eq:firstpc}
p_{(i,j)}^1 = \bm{\xi}^\top_1 \left(\bm{x}_{\left(i,j\right)} - \bar{\bm{x}} \right) \,.
\end{equation}
According to their spatial locations, all $p_{(i,j)}^1$ from an image are formed into a 2-D matrix whose dimensions are $h\times w$. We call that matrix as \textbf{{indicator matrix}}:
\begin{equation}
\label{eq:indmat}
P^1 = \left[
  			\begin{array}{cccc}  
          					p_{(1,1)}^1 & p_{(1,2)}^1 & \ldots & p_{(1,w)}^1 \\  
						p_{(2,1)}^1 & p_{(2,2)}^1 & \ldots & p_{(2,w)}^1 \\
          					\vdots & \vdots & \ddots & \vdots \\  
          					p_{(h,1)}^1 & p_{(h,2)}^1 & \ldots & p_{(h,w)}^1
 			\end{array}  
 	      \right]\, \in \mathcal{R}^{h\times w}.
\end{equation}

\begin{algorithm}[t]
\caption{Finding the largest connected component}
\label{algo:bwconn}
\begin{algorithmic}[1]{
\REQUIRE {The resized indicator matrix $P^1$ corresponding to an image $I$};
\STATE {Transform $P^1$ into a binary map $\hat{P}^1$, \\where $\hat{p}^1_{(i,j)} = \left\{
\begin{aligned}
1 & &\text{if } p^1_{(i,j)} > 0 \\
0 & & \text{otherwise}
\end{aligned}\right.$};
\STATE {Select one pixel $p$ in $\hat{P}^1$ as the starting point};
\WHILE {True}
\STATE {Use a flood-fill algorithm to label all the pixels in the connected component containing $p$};
\IF {All the pixels are labeled}
\STATE {Break};
\ENDIF
\STATE {Search for the next unlabeled pixel as $p$};
\ENDWHILE
\STATE {Obtain the connectivity of the connected components, and their corresponding size (pixel numbers)};
\STATE {Select the connected component $\hat{P}_{\rm c}^1$ with the largest pixel number};
\RETURN {The largest connected component $\hat{P}_{\rm c}^1$.}
}\end{algorithmic}
\end{algorithm}

$P^1$ contains positive (negative) values which can reflect the positive (negative) correlations of these deep descriptors. The larger the absolute value is, the higher the positive (negative) correlation will be. Because $\bm{\xi}_1$ is obtained through all $N$ images in that image set, the positive correlation could indicate the \emph{common characteristic} through $N$ images. Specifically, in the object co-localization scenario, the corresponding positive correlation indicates indeed the \emph{common object} inside these images.

Therefore, the value zero could be used as a natural threshold for dividing $P^1$ of one image into two parts: one part has positive values indicating the common object, and the other part has negative values presenting background or objects that rarely appear. Additionally, if $P^1$ of an image has no positive value, it indicates that no common object exists in that image, which can be used for detecting noisy images.

In practice, for localizing objects, $P^1$ is resized by the nearest interpolation, such that its size is the same as that of the input image. Since the nearest interpolation is the zero-order interpolation method, it will not change the signs of the numbers in $P^1$. Thus, the resized $P^1$ can be used for localizing the common object according to the aforementioned principle with the natural threshold (i.e., the value zero). Meanwhile, we employ the algorithm described in Algo.~\ref{algo:bwconn} to collect the largest connected component of the positive regions in the resized $P^1$ to remove several small noisy positive parts. Then, the minimum rectangle bounding box which contains the largest connected component of positive regions is returned as our object co-localization prediction for each image. The whole procedure of the proposed DDT method is shown in Algo.~\ref{algo:ddt}.

\begin{algorithm}[t]
\caption{Deep Descriptor Transforming (DDT)}
\label{algo:ddt}
\begin{algorithmic}[1]{
\REQUIRE {A set of $N$ images containing the common object, and a pre-trained CNN model $\mathcal{F}$};
\STATE {Feed these images with their original resolutions into $\mathcal{F}$};
\STATE {Collect the corresponding convolutional descriptors $X^1,\ldots,X^N$ from the last convolutional layer of $\mathcal{F}$};
\STATE {Calculate the mean vector $\bar{\bm{x}}$ of all the descriptors using Eq.~\ref{eq:mean}};
\STATE {Compute the covariance matrix ${\rm{Cov}} (\bm{x})$ of these deep descriptors based on Eq.~\ref{eq:cov}};
\STATE {Compute the eigenvectors $\bm{\xi}_1,\ldots, \bm{\xi}_d$ of ${\rm{Cov}} (\bm{x})$};
\STATE {Select $\bm{\xi}_1$ with the largest eigenvalue as the main transforming direction};
\REPEAT
	\STATE {Calculate the indicator matrix $P^1$ for image $I$ based on Eq.~\ref{eq:firstpc} and Eq.~\ref{eq:indmat}};
	\STATE {Resize $P^1$ into its image's resolution by nearest interpolation};
	\STATE {Collect the largest connected component $\hat{P}_{\rm c}^1$ of these positive regions of the resized $P^1$ by Algo.~\ref{algo:bwconn}};
	\STATE {Obtain the minimum rectangle bounding box covering $\hat{P}_{\rm c}^1$ as the prediction};
\UNTIL {All the $N$ images are done};
\RETURN {The minimum rectangle bounding boxes}.
}\end{algorithmic}
\end{algorithm}

\subsection{Discussions and Analyses}

In this section, we investigate the effectiveness of DDT by comparing with SCDA.

\begin{figure*}[t]
 \centering
 \includegraphics[width=\textwidth]{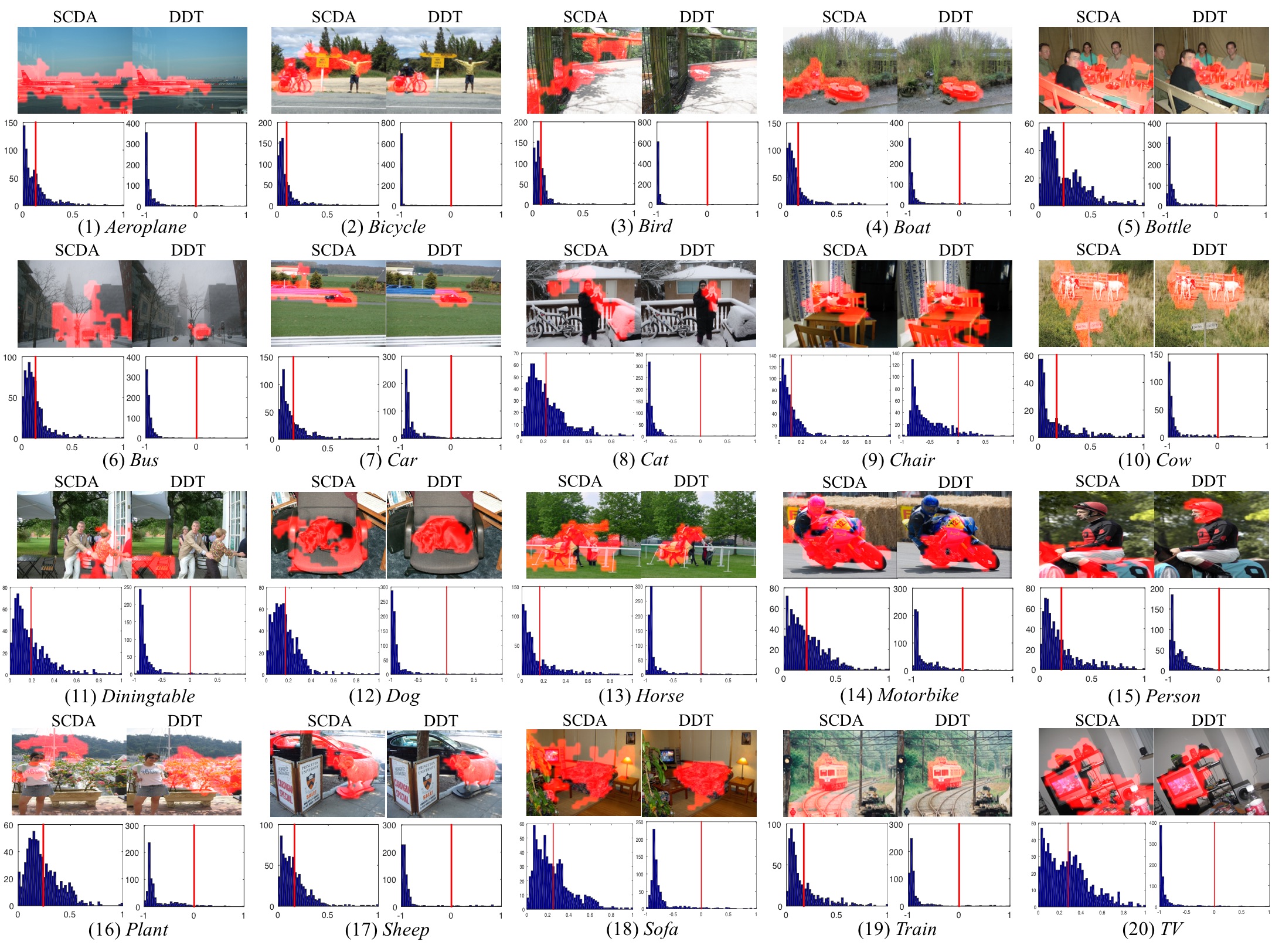}
 \caption{Examples of twenty categories from the \emph{PASCAL VOC 2007} dataset~\citep{voc2015}. The first column of each sub-figure is produced by SCDA, the second column is by our DDT. The red vertical lines in the histogram plots indicate the corresponding thresholds for localizing objects. The selected regions in images are highlighted in red. (Best viewed in color and zoomed in.)}
 \label{fig:heatmap2}
\end{figure*}

As shown in Fig.~\ref{fig:heatmap2}, the object localization regions of SCDA and DDT are highlighted in red. Because SCDA only considers the information from a single image, for example, in Fig.~\ref{fig:heatmap2}~(2),  ``bike'', ``person'' and even ``guide-board'' are all detected as main objects. Similar observations could be found in Fig.~\ref{fig:heatmap2}~(5), (13), (17), (18), etc.

Furthermore, we normalize the values (all positive) of the aggregation map of SCDA into the scale of $\left[0,1\right]$, and calculate the mean value (which is taken as the object localization threshold in SCDA). The histogram of the normalized values in aggregation map is also shown in the corresponding sub-figure in Fig.~\ref{fig:heatmap2}. The red vertical line corresponds to the threshold. We can find that, beyond the threshold, there are still many values. It gives an explanation about why SCDA highlights more regions.

Whilst, for DDT, it leverages the whole image set to transform these deep descriptors into $P^1$. Thus, for the \emph{bicycle} class (cf. Fig.~\ref{fig:heatmap2}~(2)), DDT can accurately locate the ``bicycle'' object. The histogram of DDT is also drawn. But, $P^1$ has both positive and negative values. We normalize $P^1$ into the $\left[-1,1\right]$ scale this time. Apparently, few values are larger than the DDT threshold (i.e., the value zero). More importantly, many values are close to $-1$ which indicates the strong negative correlation. This observation validates the effectiveness of DDT in image co-localization. As another example shown in Fig.~\ref{fig:heatmap2}~(11), SCDA even wrongly locates ``person'' in the image belonging to the \emph{diningtable} class. While, DDT can correctly and accurately locate the ``diningtable'' image region. More examples are presented in Fig.~\ref{fig:heatmap2}. In that figure, some failure cases can be also found, e.g., the \emph{chair} class in Fig.~\ref{fig:heatmap2}~(9).

In addition, the normalized $P^1$ can be also used as localization probability scores. Combining it with conditional random filed techniques might produce more accurate object boundaries. Thus, DDT can be modified slightly in that way, and then perform the co-segmentation problem.

\subsection{Multiple Layer Ensemble}

As is well known, CNNs are composed of multiple processing layers to learn representations of images with multiple levels of abstraction. Different layers will learn different level visual information~\citep{visualcnn2014}. Lower layers have more general representations (e.g., textures and shapes), and they can capture more detailed visual cues. By contrast, the learned representations of deeper layers contain more semantic information (i.e., high-level concepts). Thus, deeper layers are good at abstraction, but they lack visual details. Apparently, lower layer and deeper layer are complementary with each other. Based on this, several previous work, e.g., \citet{Bharath15CVPR,Jonathan15CVPR}, aggregate the information of multiple layers to boost the final performance on their computer vision tasks.

Inspired by them, we also incorporate the lower convolutional layer in pre-trained CNNs to supply finer detailed information for object co-localization, which is named as {DDT$^+$}.

\begin{figure}[t]
 \centering
 \subfloat[\emph{Bicycle}]  { \includegraphics[width=0.9\columnwidth,height=5em]{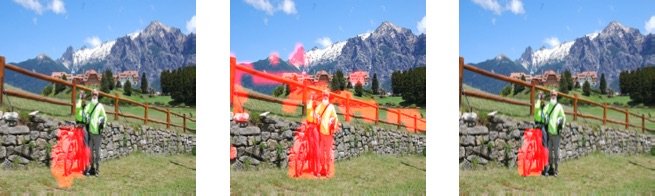} \label{fig:ddtp1} }

 \subfloat[\emph{Bottle}] { \includegraphics[width=0.9\columnwidth,height=5em]{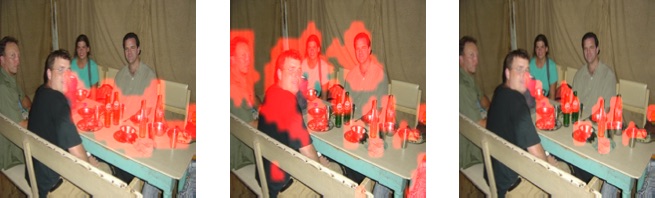} \label{fig:ddtp2} }

 \subfloat[\emph{Car}] { \includegraphics[width=0.9\columnwidth,height=5em]{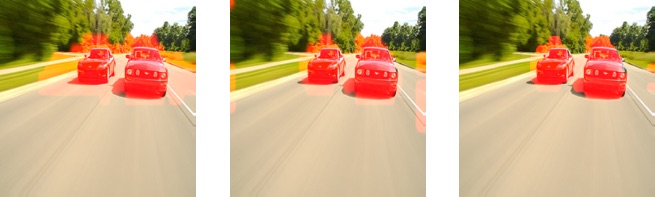} \label{fig:ddtp3} }

 \subfloat[\emph{Cow}] { \includegraphics[width=0.9\columnwidth,height=5em]{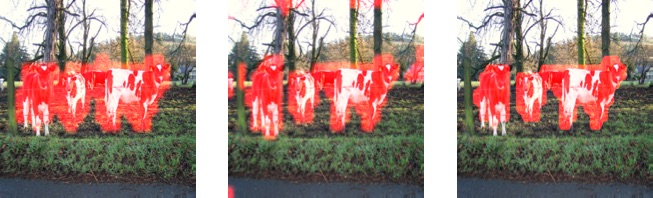} \label{fig:ddtp4} }

 \subfloat[\emph{Horse}] { \includegraphics[width=0.9\columnwidth,height=5em]{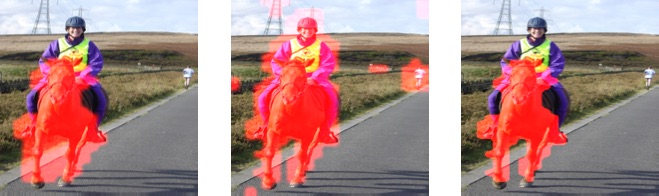} \label{fig:ddtp5} }
 \caption{Examples from five randomly sampled categories of \emph{PASCAL VOC 2007}~\citep{voc2015}. The red highlighted regions in images are  detected as containing common objects by our proposed methods. In each sub-figure, the first column presents the prediction by our DDT (cf. Algo.~\ref{algo:ddt}). The middle column shows the DDT's result based on the lower convolutional layer. The last column are the predicted results by our DDT$^+$. (Best viewed in color and zoomed in.)} \label{fig:ddtp}
\end{figure}

Concretely, as aforementioned in Algo.~\ref{algo:ddt}, we can obtain $\hat{P}_{\rm c}^1$ of the resize $P^1$ for each image from the last convolutional layer by our DDT. Several visualization examples of $\hat{P}_{\rm c}^1$ are shown in the first column of Fig.~\ref{fig:ddtp}. In DDT$^+$, beyond that, those deep descriptors from the previous convolutional layer before the last one are also used for generating its corresponding resized $P^1$, which is notated as $P^1_{\rm prev}$. For $P^1_{\rm prev}$, we directly transform it into a binary map $\hat{P}^1_{\rm prev}$. In the middle column of Fig.~\ref{fig:ddtp}, the red highlighted regions represent the co-localization results by $\hat{P}^1_{\rm prev}$. Since the activations from the previous convolutional layer are less related to the high-level semantic meaning than those from the last convolutional layer, other objects not belonging to the common object category are also being detected. However, the localization boundaries are much finer than $\hat{P}_{\rm c}^1$. Therefore, we combine $\hat{P}_{\rm c}^1$ and $\hat{P}^1_{\rm prev}$ together to obtain the final co-localization prediction as follows:
\begin{equation}
\label{eq:cap}
\hat{P}_{\rm c}^1 \cap \hat{P}^1_{\rm prev}\,.
\end{equation}
As shown in the last column of Fig.~\ref{fig:ddtp}, the co-localization visualization results of DDT$^+$ are better than the results of DDT, especially for the \emph{bottle} class. In addition, from the quantitative perspective, DDT$^+$ will bring on average 1.5\% improvements on image co-localization (cf. Table~\ref{table:voc07}, Table~\ref{table:voc12} and Table~\ref{table:unseen}).

\section{Experiments}\label{sec:experi}

In this section, we first introduce the evaluation metric and datasets used in image co-localization. Then, we compare the empirical results of our DDT and DDT$^+$ with other state-of-the-arts on these datasets. The computational cost is reported too. Moreover, the results in Sec.~\ref{sec:unseen} and Sec.~\ref{sec:noise} illustrate the generalization ability and robustness of the proposed method. Furthermore, we will discuss the ability of DDT to utilize web data as valid augmentation for improving the accuracy of traditional image recognition and object detection tasks. Finally, the further study in Sec.~\ref{sec:future} reveals DDT might deal with part-based image co-localization, which is a novel and challenging problem.

In our experiments, the images keep the original image resolutions. For the pre-trained deep model, the publicly available VGG-19 model~\citep{vgg16} is employed  to perform DDT by extracting deep convolution descriptors from the last convolution layer (i.e., the $\text{relu}_{5\_4}$ layer) and employed to perform DDT$^+$ by using both the last convolution layer (i.e., the $\text{relu}_{5\_4}$ layer) and its previous layer (i.e., the $\text{relu}_{5\_3}$ layer). We use the open-source library MatConvNet~\citep{matconvnet} for conducting experiments. All the experiments are run on a computer with Intel Xeon E5-2660 v3, 500G main memory, and a K80 GPU.

\subsection{Evaluation Metric and Datasets}

Following previous image co-localization works~\citep{yaoeccv2016,chicvpr2015,tangcvpr2014}, we take the correct localization (CorLoc) metric for evaluating the proposed method. CorLoc is defined as the percentage of images correctly localized according to the PASCAL-criterion~\citep{voc2015}:
\begin{equation}
\frac{{\rm area} (B_{\rm p} \cap B_{\rm gt})}{{\rm area} (B_{\rm p} \cup B_{\rm gt})}>0.5\,,
\end{equation}
where $B_{\rm p}$ is the predicted bounding box and $B_{\rm gt}$ is the ground-truth bounding box. All CorLoc results are reported in percentages.

Our experiments are conducted on four challenging datasets commonly used in image co-localization, i.e., the \emph{Object Discovery} dataset~\citep{rubinscvpr2013}, the \emph{PASCAL VOC 2007}/\emph{VOC 2012} dataset~\citep{voc2015} and the \emph{ImageNet Subsets} dataset \citep{yaoeccv2016}.

For experiments on the \emph{PASCAL VOC} datasets, we follow~\citet{chicvpr2015,yaoeccv2016,joulineccv2014} to use all images in the \emph{trainval} set (excluding images that only contain object instances annotated as \emph{difficult} or \emph{truncated}). For \emph{Object Discovery}, we use the 100-image subset following~\citet{rubinscvpr2013,chicvpr2015} in order to make an appropriate comparison with other methods.

In addition, \emph{Object Discovery} has 18\%, 11\% and 7\% noisy images in the \emph{Airplane}, \emph{Car} and \emph{Horse} categories, respectively. These noisy images contain no object belonging to their category, as the third image shown in Fig.~\ref{fig:pipeline}. Particularly, in Sec.~\ref{sec:noise}, we quantitatively measure the ability of our proposed DDT to identify these noisy images.

To further investigate the generalization ability of DDT, \emph{ImageNet Subsets}~\citep{yaoeccv2016} are used, which contain six subsets/categories. These subsets are held-out categories from the 1000-label ILSVRC classification~\citep{russaijcv2015}. That is to say, these subsets are ``unseen'' by pre-trained CNN models. Experimental results in Sec.~\ref{sec:unseen} show that our proposed methods is insensitive to the object category.

\begin{table}[t!]
 \caption{Comparisons of CorLoc on \emph{Object Discovery}.} \label{table:objdisc}
 \centering
 \setlength{\tabcolsep}{3pt}
 \begin{tabular}{l||c|c|c||c}
  \hline
  {Methods}  & {\emph{airplane}} & {\emph{car}} & {\emph{horse}} & {\textbf{Mean}} \\  
  \hline
  \citet{joulincoscvpr2010}  & 32.93  & 66.29 &  54.84 &  51.35 \\
  \citet{joulinmccvpr2010}  & 57.32  & 64.04 &  52.69  & 58.02\\
  \citet{rubinscvpr2013}  & 74.39  & 87.64  & 63.44 &  75.16\\
  \citet{tangcvpr2014}  & 71.95 &  93.26  & 64.52  & 76.58\\
  SCDA & 87.80  &  	86.52  &  	75.37  &  	83.20 \\
  \citet{chicvpr2015}  & 82.93  & 94.38  & 75.27 &  84.19\\
  \hline
  Our DDT & \textbf{91.46}	 & \textbf{95.51}	 & \textbf{77.42}	 & \textbf{88.13}\\
  Our DDT$^+$ & \textbf{91.46}	 & {94.38}	 & {76.34}	 & {87.39}\\
  \hline
 \end{tabular}
\end{table}

\begin{table*}[th!]
 \caption{Comparisons of the CorLoc metric with state-of-the-art co-localization methods on \emph{VOC 2007}.} \label{table:voc07}
 \centering
 \setlength{\tabcolsep}{0pt}
\resizebox{\textwidth}{!}{
 \begin{tabular}{l||c|c|c|c|c|c|c|c|c|c|c|c|c|c|c|c|c|c|c|c||c}
  \hline
  {Methods} & \emph{aero} & \emph{bike} & \emph{bird}& \emph{boat}& \emph{bottle}& \emph{bus}& \emph{car}& \emph{cat}& \emph{chair}& \emph{cow}& \emph{table}& \emph{dog}& \emph{horse}& \emph{mbike}& \emph{person}& \emph{plant}& \emph{sheep}& \emph{sofa}& \emph{train}& \emph{tv} & \textbf{Mean}\\  
  \hline
  \citet{joulineccv2014}  & 32.8  & 17.3  & 20.9 &  18.2  & 4.5  & 26.9 &  32.7  & 41.0 &  5.8  & 29.1  & \textbf{34.5}  & 31.6  & 26.1  & 40.4  & 17.9  & 11.8 &  25.0  & 27.5  & 35.6  & 12.1  & 24.6  \\
  SCDA & 54.4   & 	27.2 	  & 43.4 	  & 13.5   & 	2.8   & 	39.3 	  & 44.5   & 	48.0   & 	6.2 	  & 32.0   & 	16.3   & 	49.8 	  & 51.5   & 	49.7   & 	7.7   & 	6.1  &  	22.1  &  	22.6 	  & 46.4   & 	6.1   & 	29.5 \\
  \citet{chicvpr2015}   & 50.3 &  42.8 &  30.0  & 18.5  & 4.0  & 62.3  & \textbf{64.5} &  42.5  & 8.6  & \textbf{49.0}  & 12.2  & 44.0  & 64.1  & 57.2  & 15.3  & 9.4  & 30.9  & 34.0 &  61.6  & {31.5}  & 36.6\\
  \citet{yaoeccv2016}   & \textbf{73.1}  & 45.0 &  43.4  & \textbf{27.7} &  6.8  & 53.3  & 58.3  & 45.0  & 6.2  & 48.0  & 14.3  & 47.3  & 69.4  & 66.8 &  \textbf{24.3}  & 12.8  & \textbf{51.5}  & 25.5  & 65.2  & 16.8  & 40.0 \\
  \hline
  Our DDT  & 67.3 	& {63.3}  & 	{61.3}  & 	22.7  & 	\textbf{8.5} 	 & \textbf{64.8}  & 	57.0  & 	\textbf{80.5} 	 & {9.4} 	 & \textbf{49.0} 	 & 22.5 	 & \textbf{72.6} &  	{73.8}  & 	\textbf{69.0}  & 	7.2  & 	\textbf{15.0}  & 	35.3  & 	{54.7} 	 & \textbf{75.0} 	 & 29.4 	 & {46.9} \\
  Our DDT$^+$  &  71.4 	 & \textbf{65.6}  & 	\textbf{64.6}  & 	25.5 	 & \textbf{8.5} 	 & \textbf{64.8}  & 	61.3  & 	\textbf{80.5} 	 & \textbf{10.3}  & 	\textbf{49.0}  & 	26.5  & 	\textbf{72.6}  & 	\textbf{75.2}  & 	\textbf{69.0}  & 	9.9 	 & 12.2 	 & 39.7  & 	\textbf{55.7} &  	\textbf{75.0} 	 & \textbf{32.5} 	 & \textbf{48.5}  \\
  \hline
 \end{tabular}
}
\end{table*}

\begin{table*}[th!]
 \caption{Comparisons of the CorLoc metric with state-of-the-art co-localization methods on \emph{VOC 2012}.} \label{table:voc12}
 \centering
 \setlength{\tabcolsep}{0pt}
\resizebox{\textwidth}{!}{
 \begin{tabular}{l||c|c|c|c|c|c|c|c|c|c|c|c|c|c|c|c|c|c|c|c||c}
  \hline
  {Methods} & \emph{aero} & \emph{bike} & \emph{bird}& \emph{boat}& \emph{bottle}& \emph{bus}& \emph{car}& \emph{cat}& \emph{chair}& \emph{cow}& \emph{table}& \emph{dog}& \emph{horse}& \emph{mbike}& \emph{person}& \emph{plant}& \emph{sheep}& \emph{sofa}& \emph{train}& \emph{tv} & \textbf{Mean}\\  
  \hline
  SCDA & 60.8  & 	41.7 	 & 38.6 	 & 21.8  & 	7.4  & 	67.6  & 	38.8 	 & 57.4  & 	16.0  & 	34.0  &  	23.9  & 	53.8  & 	47.3  & 	54.8 &  	7.9  & 	9.9  & 	25.3  & 	23.2  & 	50.2  & 	10.1  & 	34.5  \\
  \citet{chicvpr2015}  & 57.0 	 & 41.2  & 	36.0  & 	26.9  & 	5.0 	 & 81.1 	 & \textbf{54.6}  & 	50.9  & 	18.2  & 	54.0  & 	\textbf{31.2}  & 	44.9  & 	61.8  & 	48.0 &  	13.0 	 & 11.7  & 	51.4  & 	45.3  & 	64.6  & 	\textbf{39.2}  & 	41.8   \\
  \citet{yaoeccv2016}   &65.7 	 & 57.8  & 	47.9 	 & 28.9 	 & 6.0 	 & 74.9 	 & 48.4  & 	48.4  & 	14.6  & 	\textbf{54.4}  & 	23.9  & 	50.2  & 	\textbf{69.9}  & 	68.4  & 	\textbf{24.0}  & 	14.2  & 	\textbf{52.7} &  	30.9  & 	72.4 	 & 21.6 	 & 43.8  \\
  \hline
  Our DDT  & {76.7}  & 	{67.1} 	 & {57.9} 	 & {30.5}  & 	{13.0}  & 	{81.9}  & 	48.3  & 	\textbf{75.7} 	 & {18.4}  & 	48.8  & 	27.5 	 & {71.8}  & 	66.8 &  	{73.7} 	 & 6.1  & 	\textbf{18.5}  & 	38.0  & 	{54.7}  & 	\textbf{78.6}  & 	34.6 	 & {49.4}  \\
  Our DDT$^+$  &  \textbf{77.9}  & 	\textbf{67.7} 	 & \textbf{61.8} 	 & \textbf{33.8}  & 	\textbf{14.2}  & 	\textbf{82.5} 	 & 53.0 	 & 75.2  & 	\textbf{18.9}  & 	53.5  & 	28.3 	 & \textbf{73.8} 	 & 68.7 	 & \textbf{77.5}  & 	8.4  & 	17.6  & 	40.8  & 	\textbf{55.3}  & 	\textbf{78.6}  & 	35.0  & 	\textbf{51.1}  \\
  \hline
 \end{tabular}
}
\end{table*}

\subsection{Comparisons with State-of-the-Arts}

In this section, we compare the image co-localization performance of our methods with state-of-the-art methods including both image co-localization and weakly supervised object localization.

\subsubsection{Comparisons to Image Co-Localization Methods}

We first compare the results of DDT to state-of-the-arts (including SCDA) on \emph{Object Discovery} in Table~\ref{table:objdisc}. For SCDA, we also use VGG-19 to extract the convolution descriptors and perform experiments. As shown in that table, DDT outperforms other methods by about 4\% in the mean CorLoc metric. Especially for the \emph{airplane} class, it is about 10\% higher than that of~\citet{chicvpr2015}. In addition, note that the images of each category in this dataset contain only one object, thus, SCDA can perform well. But, our DDT$^+$ gets a slightly lower CorLoc score than DDT, which is an exception in all the image co-localization datasets. In fact, for \emph{car} and \emph{horse} of the \emph{Object Discovery} dataset, DDT$^+$ only returns one more wrong prediction than DDT for each category.

For \emph{PASCAL VOC 2007} and \emph{2012}, these datasets contain diverse objects per image, which is more challenging than \emph{Object Discovery}. The comparisons of the CorLoc metric on these two datasets are reported in Table~\ref{table:voc07} and Table~\ref{table:voc12}, respectively. It is clear that on average our DDT and DDT$^+$ outperform the previous state-of-the-arts (based on deep learning) by a large margin on both datasets. Moreover, our methods work well on localizing small common objects, e.g., ``bottle'' and ``chair''. In addition, because most images of these datasets have multiple objects, which do not obey SCDA's assumption, SCDA performs poorly in the complicated environment. For fair comparisons, we also use VGG-19 to extract the fully connected representations of the object proposals in~\citep{yaoeccv2016}, and then perform the remaining processes of their method (the source codes are provided by the authors). As aforementioned, due to the high dependence on the quality of object proposals, their mean CorLoc metric of VGG-19 is 41.9\% and 45.6\% on \emph{VOC 2007} and \emph{2012}, respectively. The improvements are limited, and the performance is still significantly worse than ours.

\subsubsection{Comparisons to Weakly Supervised Localization Methods}

To further verify the effectiveness of our methods, we also compare DDT and DDT$^+$ with some state-of-the-art methods for weakly supervised object localization. Table~\ref{table:voc07weak} illustrates these empirical results on \emph{VOC 2007}. Particularly, DDT achieves 46.9\% on average which is higher than most WSOL methods in the literature. DDT$^+$ achieves 48.5\% on average, and it even performs better than the state-of-the-art in WSOL (i.e., \citet{wangeccv2014}) which is also a deep learning based approach. Meanwhile, note that our methods do \emph{not} use any negative data for co-localization. Moreover, our methods could handle noisy data (cf. Sec.~\ref{sec:noise}). But, existing WSOL methods are not designed to deal with noise.

\begin{table*}[th!]
 \caption{Comparisons of the CorLoc metric with weakly supervised object localization methods on \emph{VOC 2007}. Note that, the ``$\checkmark$'' in the ``Neg.'' column indicates that these WSOL methods require access to a negative image set, whereas our DDT does not.} \label{table:voc07weak}
 \centering
 \setlength{\tabcolsep}{0pt}
\resizebox{\textwidth}{!}{
 \begin{tabular}{l|c||c|c|c|c|c|c|c|c|c|c|c|c|c|c|c|c|c|c|c|c||c}
  \hline
  {Methods}  & Neg. & \emph{aero} & \emph{bike} & \emph{bird}& \emph{boat}& \emph{bottle}& \emph{bus}& \emph{car}& \emph{cat}& \emph{chair}& \emph{cow}& \emph{table}& \emph{dog}& \emph{horse}& \emph{mbike}& \emph{person}& \emph{plant}& \emph{sheep}& \emph{sofa}& \emph{train}& \emph{tv} & \textbf{Mean}\\  
  \hline
 \citet{sivacvpr2011} &$\checkmark$ & 42.4  & 	46.5 	 & 18.2  & 	8.8 	 & 2.9 	 & 40.9 	 & 73.2 	 & 44.8  & 	5.4  & 	30.5 &  	19.0 	 & 34.0 	 & 48.8  & 	65.3  & 	8.2 	 & 9.4  & 	16.7  & 	32.3  & 	54.8  & 	5.5  & 	30.4 \\
  \citet{shiiccv2013} &$\checkmark$ & 67.3  & 	54.4 	 & 34.3  & 	17.8  & 	1.3 	 & 46.6 	 & 60.7 	 & 68.9  & 	2.5 	 & 32.4 	 & 16.2 	 & 58.9 	 & 51.5  & 	64.6  & 	18.2 	 & 3.1  & 	20.9 	 & 34.7 	 & 63.4 	 & 5.9  & 	36.2  \\
  \citet{cinbiscvpr2014} &$\checkmark$ & 56.6 	 & 58.3 &  	28.4  & 	20.7  & 	6.8 	 & 54.9  & 	69.1 &  	20.8 	 & 9.2  & 	50.5 	 & 10.2  & 	29.0 &  	58.0 	 & 64.9  & 	36.7  & 	18.7  & 	56.5  & 	13.2  & 	54.9  & 	59.4 	 & 38.8  \\
  \citet{wangiccv2015} & $\checkmark$& 37.7 	 & 58.8 &  	39.0 &  	4.7  & 	4.0 	 & 48.4 	 & 70.0 	 & 63.7 	 & 9.0  & 	54.2  & 	\textbf{33.3} 	 & 37.4 	 & 61.6  & 	57.6 &  	30.1 	 & 31.7  & 	32.4 &  	52.8  & 	49.0  & 	27.8 	 & 40.2 \\
  \citet{bilencvpr2015} &$\checkmark$ & 66.4  & 	59.3  & 	42.7 	 & 20.4 	 & \textbf{21.3} 	 & 63.4 	 & \textbf{74.3} 	 & 59.6  & 	21.1  & 	58.2 	 & 14.0 	 & 38.5 	 & 49.5  & 	60.0  & 	19.8 	 & 39.2  & 	41.7 &  	30.1 	 & 50.2  & 	44.1  & 	43.7  \\
  \citet{renpami2016} &$\checkmark$ & 79.2 	 & 56.9  & 	46.0  & 	12.2  & 	15.7  & 	58.4  & 	71.4  & 	48.6  & 	7.2 	 & \textbf{69.9}  & 	16.7 &  	47.4  & 	44.2  & 	\textbf{75.5} 	 & \textbf{41.2} 	 & \textbf{39.6}  & 	47.4  & 	32.2  & 	49.8  & 	18.6  & 	43.9 \\
 \citet{wangeccv2014} &$\checkmark$ & \textbf{80.1} 	 & {63.9} 	 & 51.5 &  	14.9  & 	21.0  & 	55.7  & 	74.2 &  	43.5  & 	\textbf{26.2}  & 	53.4  & 	16.3  & 	56.7  & 	58.3 	 & 69.5 &  	14.1 	 & 38.3  & 	\textbf{58.8}  & 	47.2  & 	49.1  & 	\textbf{60.9}  & 	{47.7} \\
  \hline
  Our DDT &  & 67.3 	& 63.3  & 	{61.3}  & 	{22.7}  & 	8.5 	 & \textbf{64.8}  & 	57.0  & 	\textbf{80.5} 	 & 9.4 	 & 49.0 	 & 22.5 	 & \textbf{72.6} &  	{73.8}  & 	69.0  & 	7.2  & 	15.0  & 	35.3  & 	{54.7} 	 & \textbf{75.0} 	 & 29.4 	 & 46.9 \\
  Our DDT$^+$  &  & 71.4 	 & \textbf{65.6}  & 	\textbf{64.6}  & 	\textbf{25.5} 	 & 8.5 	 & \textbf{64.8}  & 	61.3  & 	\textbf{80.5} 	 & 10.3  & 	49.0  & 	26.5  & 	\textbf{72.6}  & 	\textbf{75.2}  & 	69.0  & 	9.9 	 & 12.2 	 & 39.7  & 	\textbf{55.7} &  	\textbf{75.0} 	 & 32.5 	 & \textbf{48.5}  \\
  \hline
 \end{tabular}
}
\end{table*}

\subsection{Computational Costs of DDT/DDT$^+$}

Here, we take the total 171 images in the \emph{aeroplane} category of \emph{VOC 2007} as examples to report the computational costs. The average image resolution of the 171 images is $350\times 498$. The computational time of DDT has two main components: one is for feature extraction, the other is for deep descriptor transforming (cf. Algo.~\ref{algo:ddt}). Because we just need the first principal component, the transforming time on all the 120,941 descriptors of 512-d is only 5.7 seconds. The average descriptor extraction time is 0.18 second/image on GPU and 0.86 second/image on CPU, respectively. For DDT$^+$, it has the same deep descriptor extraction time. Although it needs descriptors from two convolutional layers, it only requires one time feed-forward processing. The deep descriptor transforming time of DDT$^+$ is only 11.9 seconds for these 171 images. These numbers above could ensure the efficiency of the proposed methods in real-world applications.

\subsection{Unseen Classes Apart from \emph{ImageNet}}\label{sec:unseen}

In order to justify the generalization ability of the proposed methods, we also conduct experiments on some images (of six subsets) disjoint with the images from \emph{ImageNet}. Note that, the six categories (i.e., ``chipmunk'', ``rhino'', ``stoat'', ``racoon'', ``rake'' and ``wheelchair'') of these images are unseen by pre-trained models. The six subsets were provided in~\citep{yaoeccv2016}. Table~\ref{table:unseen} presents the CorLoc metric on these subsets. Our DDT (69.1\% on average) and DDT$^+$ (70.4\% on average) still significantly outperform other methods on all categories, especially for some difficult objects categories, e.g., \emph{rake} and \emph{wheelchair}. In addition, the mean CorLoc metric of \citep{yaoeccv2016} based on VGG-19 is only 51.6\% on this dataset.

\begin{table}[t!]
 \caption{Comparisons of the CorLoc metric with state-of-the-arts on image sets disjoint with \emph{ImageNet}.} \label{table:unseen}
 \centering
 \setlength{\tabcolsep}{0pt}
 \begin{tabular}{l||c|c|c|c|c|c||c}
  \hline
  {Methods}  & {\emph{chipm.}} & {\emph{rhino}} & {\emph{stoat}} & {\emph{racoon}} & \emph{rake} & \emph{wheelc.} & \textbf{Mean} \\  
  \hline
  \citet{chicvpr2015}  & 26.6 & 81.8 & 44.2 & 30.1 & 8.3 & 35.3& 37.7 \\
  SCDA & 32.3  & 	71.6  & 	52.9  & 	34.0 	 & 7.6  & 	28.3 	 & 37.8 \\
  \citet{yaoeccv2016} & 44.9 & 81.8& 67.3& 41.8& 14.5& 39.3  & 48.3\\
  \hline
  Our DDT & {70.3}  & 	\textbf{93.2}  & 	\textbf{80.8} 	 & {71.8}  & 	\textbf{30.3}  & 	{68.2} & 	{69.1} \\
  Our DDT$^+$ & \textbf{72.8}  & 	\textbf{93.2}  & 	\textbf{80.8} 	 & \textbf{75.7}  & 	{28.3}  & 	\textbf{71.7} & 	\textbf{70.4} \\
  \hline
 \end{tabular}
\end{table}

\begin{figure*}[t]
 \centering
 \subfloat[\emph{Chipmunk}]  { \includegraphics[width=0.45\textwidth]{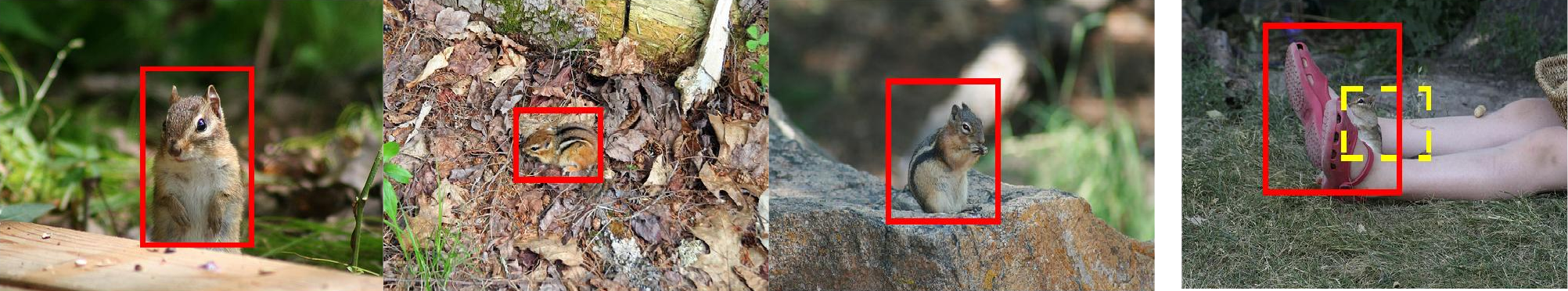} \label{fig:im1} }
	~~
 \subfloat[\emph{Rhino}] { \includegraphics[width=0.45\textwidth]{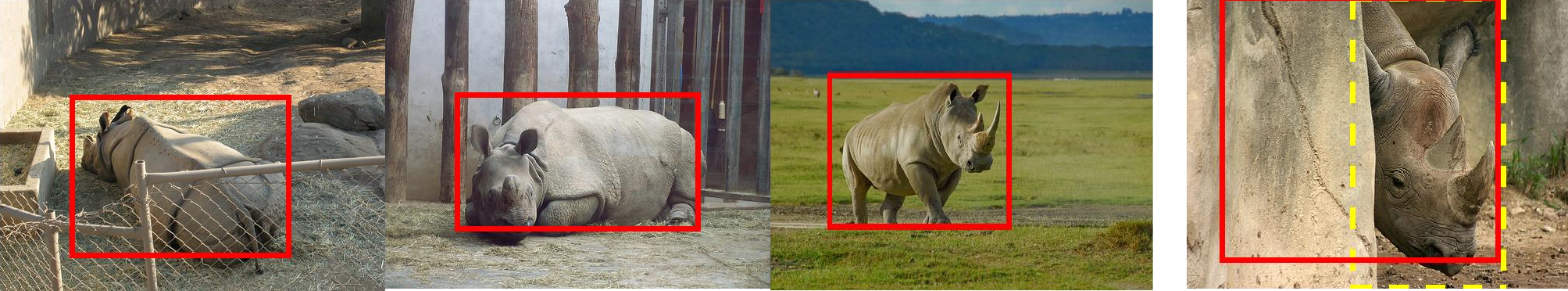} \label{fig:im2} }

 \subfloat[\emph{Stoat}] { \includegraphics[width=0.45\textwidth]{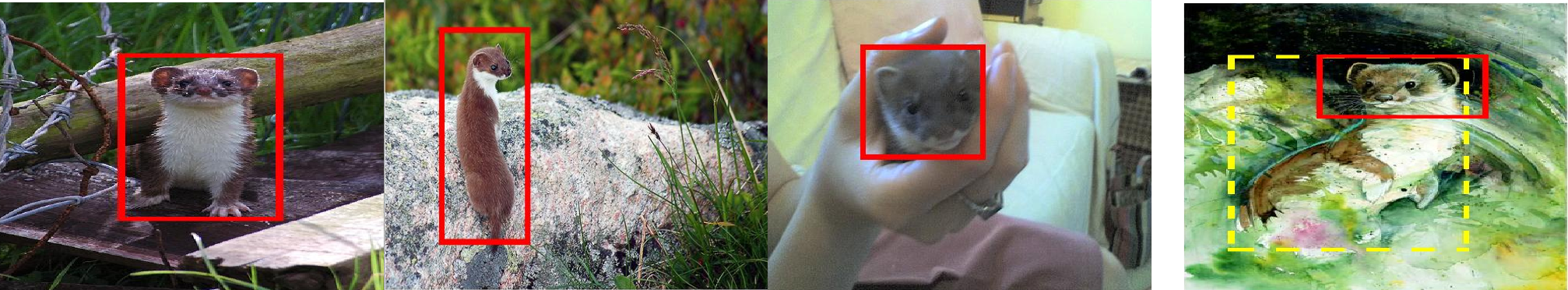} \label{fig:im3} }
	~~
 \subfloat[\emph{Racoon}] { \includegraphics[width=0.45\textwidth]{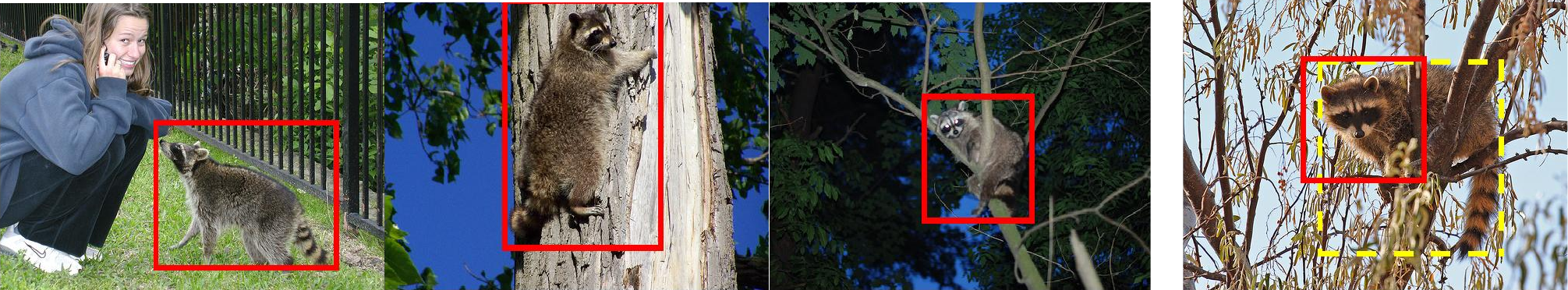} \label{fig:im4} }

 \subfloat[\emph{Rake}] { \includegraphics[width=0.45\textwidth]{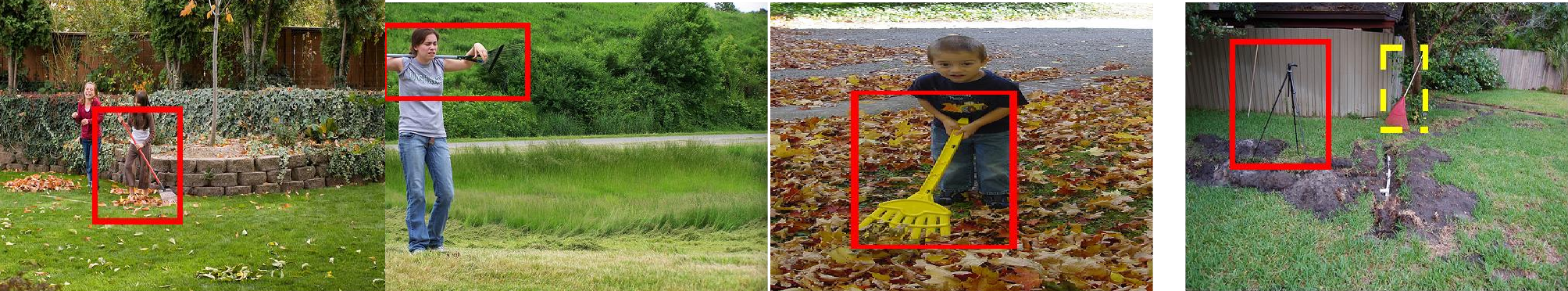} \label{fig:im5} }
	~~
 \subfloat[\emph{Wheelchair}] { \includegraphics[width=0.45\textwidth]{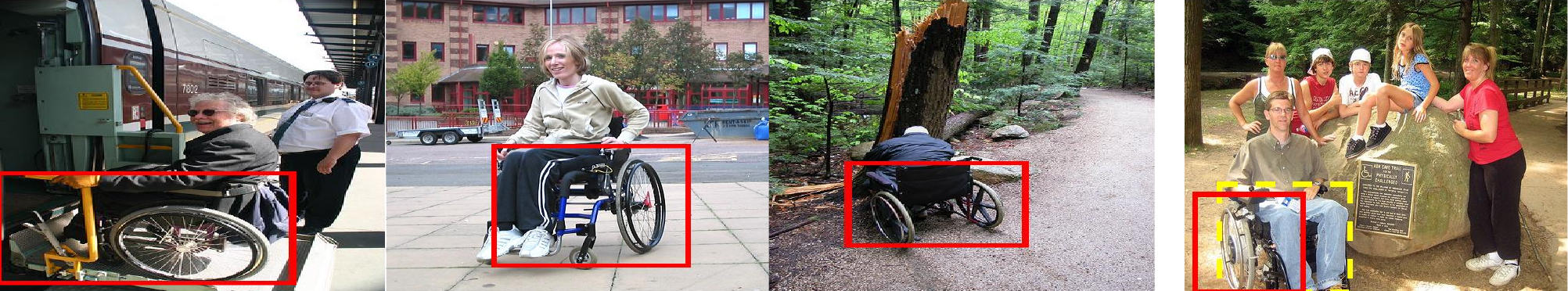} \label{fig:im6} }
 \caption{Random samples of predicted object co-localization bounding box on \emph{ImageNet Subsets}. Each sub-figure contains three successful predictions and one failure case. In these images, the red rectangle is the prediction by DDT, and the yellow dashed rectangle is the ground truth bounding box. In the successful predictions, the yellow rectangles are omitted since they are exactly the same as the red predictions. (Best viewed in color and zoomed in.)} \label{fig:bbox}
\end{figure*}

Furthermore, in Fig.~\ref{fig:bbox}, several successful predictions by DDT and also some failure cases on this dataset are provided. In particular, for ``rake'' (``wheelchair''), even though a large portion of images in these two categories contain both people and rakes (wheelchairs), our DDT could still accurately locate the common object in all the images, i.e., rakes (wheelchairs), and ignore people. This observation validates the effectiveness (especially for the high CorLoc metric on \emph{rake} and \emph{wheelchair}) of our method from the qualitative perspective.

\subsection{Detecting Noisy Images}\label{sec:noise}

\begin{figure}[t]
 \centering
 \includegraphics[width=\columnwidth]{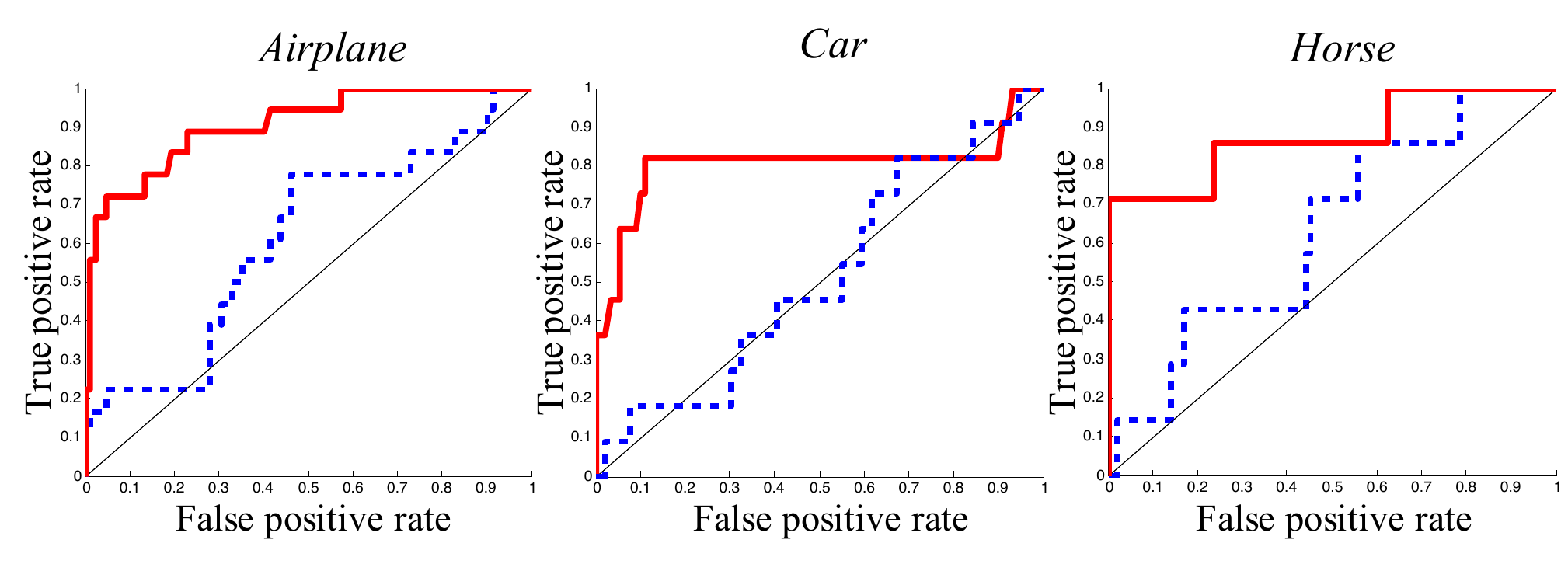}
 \caption{ROC curves illustrating the effectiveness of our DDT at identifying noisy images on the \emph{Object Discovery} dataset. The curves in red line are the ROC curves of DDT. The curves in blue dashed line present the method in~\citet{tangcvpr2014}.}
 \label{fig:roc}
\end{figure}

In this section, we quantitatively present the ability of the proposed DDT method to identify noisy images. As aforementioned, in \emph{Object Discovery}, there are 18\%, 11\% and 7\% noisy images in the corresponding categories. In our DDT, the number of positive values in $P^1$ can be interpreted as a detection score. The lower the number is, the higher the probability of noisy images will be. In particular, no positive value at all in $P^1$ presents the image as definitely a noisy image. For each category in that dataset, the ROC curve is shown in Fig.~\ref{fig:roc}, which measures how the methods correctly detect noisy images. In the literature, only the method in~\citep{tangcvpr2014} (i.e., the \texttt{Image-Box} model in that paper) could solve image co-localization with noisy data. From these figures, it is apparent to see that, in image co-localization, our DDT has significantly better performance in detecting noisy images than \texttt{Image-Box} (whose noisy detection results are obtained by re-running the publicly available code released by the authors). Meanwhile, our mean CorLoc metric without noise is about 12\% higher than theirs on \emph{Object Discovery}, cf. Table~\ref{table:objdisc}.

\subsection{DDT Augmentation based on Web Images}\label{sec:web}

As validated by previous experiments, DDT can accurately detect noisy images and meanwhile supply object bounding boxes of images (except for noisy images). Therefore, we can use DDT to process web images. In this section, we report the results of both image classification and object detection when using DDT as a tool for generating valid external data sources from free but noisy web data. This DDT based strategy is denoted as DDT augmentation.

\subsubsection{Webly-Supervised Classification}

For web based image classification, we compare DDT augmentation with the current state-of-the-art webly-supervised classification method proposed by \citet{bohan2017}. As discussed in the related work, \citet{bohan2017} proposed a group attention framework for handling web data. In their method, it employed two level attentions: the first level is designed as the group attention for filtering out noise, and the second level attention is based on the single image for capturing discriminative regions of each image.

In the experiments, we test the methods on the \emph{WebCars} and \emph{WebImageNet} datasets which are also proposed by \citet{bohan2017}. In \emph{WebCars}, there are 213,072 car images of totally 431 car model categories collected from web. In \emph{WebImageNet}, \citet{bohan2017} used 100 sub-categories of the original \emph{ImageNet} as the categories of their \emph{WebImageNet} dataset. There are 61,639 images belonging to the 100 sub-categories from web in total.

In our DDT augmentation, as what we do in Sec.~\ref{sec:noise}, we first use DDT to obtain the number of positive values in $P^1$ as the detection score for each image in every category. Here, we divide the detection score by the total number of values in $P^1$ as the noise rate which is in the range of $\left[0,1\right]$. The more the noise rate is close to zero, the higher the probability of noisy images will be. In the following, we conduct experiments with two thresholds (i.e., 0 or 0.1) with respect to the noise rate. If the noise rate of an image equals to or is smaller than the threshold, that image will be regarded as a noisy image. Then, we remove it from the original webly dataset. After doing the above processing for every category, we can obtain a relatively clean training dataset. Finally, we train deep CNN networks on that clean dataset. The other specific experimental settings of these two webly datasets follow \citet{bohan2017}.

Two kinds of deep CNN networks are conducted as the test bed for evaluating the classification performance on both two webly datasets:
\begin{itemize}
\item ``GAP'' represents the CNN model with \underline{G}lobal \underline{A}verage \underline{P}ooling as its last layer before the classification layer (i.e., fc+sigmoid), which is commonly used for the image classification task, e.g., \citet{szegedy2015going} and \citet{kaiming15residual}.
\item ``Attention'' represents the CNN model with the attention mechanism on the single image level. Because the method proposed in \citet{bohan2017} is equipped with the single image attention strategy, we also compare our method based on this baseline model for fair comparisons.
\end{itemize}

\begin{figure}[t]
 \centering
 \includegraphics[width=0.98\columnwidth,height=12em]{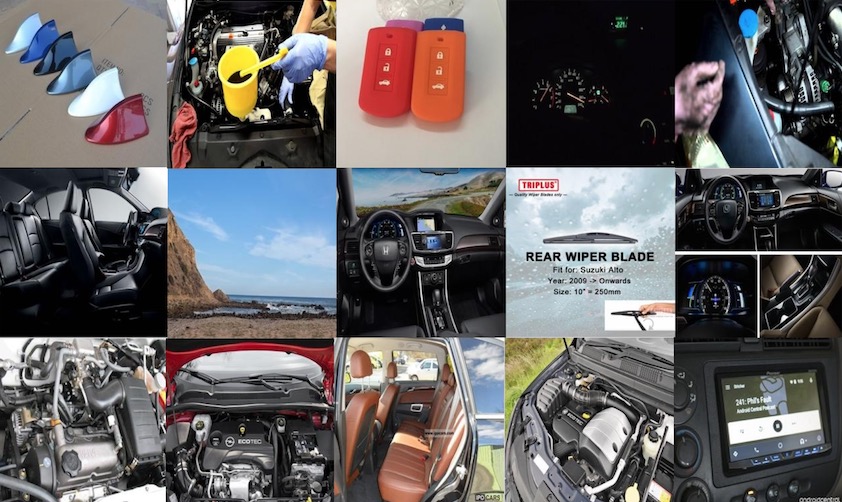}
 \caption{Examples of noisy images in the \emph{WebCars} dataset recognized by DDT.}
 \label{fig:carnoise}
\end{figure}

The quantitative comparisons of our DDT augmentation with \citet{bohan2017} are shown in Table~\ref{table:webcar} and Table~\ref{table:webimagenet}. In these tables, for example, ``DDT $\rightarrow$ GAP'' denotes that we first deploy DDT augmentation and then use the GAP model to conduct classification. As shown in these two tables, for both two base models (i.e., ``GAP'' and ``Attention''), our DDT augmentation with 0.1 threshold performs better than DDT augmentation with 0 threshold, which is reasonable. Because in many cases, the noisy images still contains several related concept regions, these (small) regions might be detected as a part of common objects. Therefore, if we set the threshold as 0.1, this kind of noisy images will be omitted. It will bring more satisfactory classification accuracy. Several detected noisy images by DDT of \emph{WebCars} are listed in Fig.~\ref{fig:carnoise}.

Comparing with the state-of-the-art (i.e., \citet{bohan2017}), our DDT augmentation with 0.1 threshold outperforms it and the GAP baseline apparently, which validate the generalization ability and the effectiveness of the proposed DDT in real-life computer vision tasks, i.e., DDT augmentation in webly-supervised classification. Meanwhile, our DDT method is easy to implement and has low computational cost, which ensures its scalability and usability in the real-world scenarios.

\begin{table}[t!]
 \caption{Comparisons of webly-supervised classification on \emph{WebCars}~\citep{bohan2017}.} \label{table:webcar}
 \centering
 \begin{tabular}{l|c||c}
  \hline
  {Methods}  & Strategy & {{Accuracy}} \\  
  \hline
  Simple-CNN & GAP & 66.86 \\
  \citet{bohan2017}  & Attention & 76.58  \\
  \hline
  Ours (thr=0) & DDT $\rightarrow$ GAP & 69.79\\
  Ours (thr=0) & DDT $\rightarrow$ Attention & {76.18} \\
  Ours (thr=0.1) & DDT $\rightarrow$ GAP & 71.66\\
  Ours (thr=0.1) & DDT $\rightarrow$ Attention & \textbf{78.92} \\
  \hline
 \end{tabular}
\end{table}

\begin{table}[t!]
 \caption{Comparisons of webly-supervised classification on \emph{WebImageNet}~\citep{bohan2017}.} \label{table:webimagenet}
 \centering
 \begin{tabular}{l|c||c}
  \hline
  {Methods}  & Strategy & {{Accuracy}} \\  
  \hline
  Simple-CNN & GAP & 58.81 \\
  \citet{bohan2017}  & Attention+Neg\footnotemark[1] & 71.24  \\
  \hline
  Ours (thr=0) & DDT $\rightarrow$ GAP & 62.31\\
  Ours (thr=0) & DDT $\rightarrow$ Attention & {69.50} \\
  Ours (thr=0.1) & DDT $\rightarrow$ GAP & 65.59\\
  Ours (thr=0.1) & DDT $\rightarrow$ Attention	 & \textbf{73.06} \\
  \hline
 \end{tabular}
\footnotemark[1]{\scriptsize{In the experiments on \emph{WebImageNet} of \citet{bohan2017}, beyond attention, they also incorporated 5,000 negative class web images for reducing noise. However, we do not require any negative images.}}
\end{table}

\subsubsection{Webly-Supervised Detection}

\begin{figure*}[t]	
 \centering
 \includegraphics[width=\textwidth, height=20em]{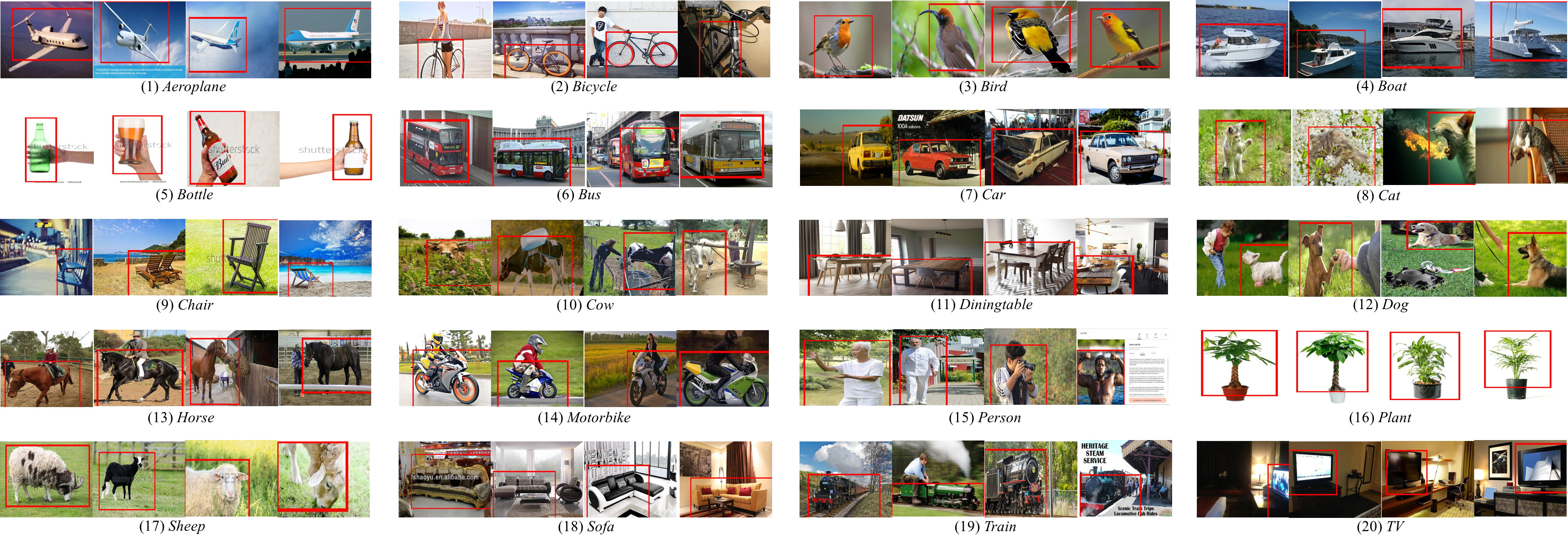}
 \caption{Examples of our \emph{WebVOC} detection dataset. The red bounding boxes in these figures are automatically labeled by the proposed DDT method. (Best viewed in color and zoomed in.)}
 \label{fig:ddt_detect}
\end{figure*}

For web based object detection, we first collect an external dataset from the Internet by Google image search engine, named \emph{WebVOC}, using the categories of the \emph{PASCAL VOC} dataset~\citep{voc2015}. In total, we collect 12,776 noisy web images, which has a similar scale as the original \emph{PASCAL VOC} dataset. As the results shown in webly-supervised classification, DDT with 0.1 threshold could be the optimal option for webly noisy images. Firstly, we also use DDT with 0.1 threshold to remove the noisy images for the images belonging to 20 categories in \emph{WebVOC}. Then, 10,081 images are remaining as valid images. Furthermore, DDT are used to automatically generate the corresponding object bounding box for each image. The generated bounding boxes by DDT are regarded as the object ``ground truth'' bounding boxes for our \emph{WebVOC} detection dataset. Several random samples of our \emph{WebVOC} dataset with the corresponding DDT generating bounding boxes are shown in Fig.~\ref{fig:ddt_detect}.

\begin{table*}[t!]
 \caption{Comparisons of detection results on the \emph{VOC 2007} test set. Note that, ``07+12'' presents the training data is the union set of VOC 2007 trainval and VOC 2012 trainval. ``COCO'' denotes that the COCO trainval set is used for training. ``DDT'' denotes that the webly data processed by DDT augmentation is used for training.} \label{table:webdetect07}
 \centering
 \setlength{\tabcolsep}{0.2pt}
 \begin{tabular}{l||c|c|c|c|c|c|c|c|c|c|c|c|c|c|c|c|c|c|c|c||c}
  \hline
  {Data}  & \emph{aero} & \emph{bike} & \emph{bird}& \emph{boat}& \emph{bottle}& \emph{bus}& \emph{car}& \emph{cat}& \emph{chair}& \emph{cow}& \emph{table}& \emph{dog}& \emph{horse}& \emph{mbike}& \emph{person}& \emph{plant}& \emph{sheep}& \emph{sofa}& \emph{train}& \emph{tv} & mAP (\%) \\  
  \hline
  07+12 & 76.5  & 79.0  & 70.9  & 65.5  & 52.1  & 83.1  & 84.7  & 86.4  & 52.0  & 81.9  & 65.7  & 84.8  & 84.6  & 77.5  & 76.7  & 38.8  & 73.6  & 73.9  & 83.0  & 72.6  & 73.2\\
  COCO\footnotemark[1]+07+12 & 84.3  & 82.0  & 77.7  & 68.9  & 65.7  & 88.1  & 88.4  & 88.9  & 63.6  & 86.3  & 70.8  & 85.9  & 87.6  & 80.1  & 82.3  & 53.6  & 80.4  & 75.8  & 86.6  & 78.9  & 78.8 \\
  \hline
  DDT+07+12 & 77.6 & 	82.2 & 	77.2 & 	64.9 & 	61.2 & 	85.4 	& 87.2 	& 88.6 & 	58.2 & 	82.6 & 	69.7 & 	85.9 & 	87.0 & 	78.9 & 	78.5 & 	46.3 	& 76.6 & 	73.5 	& 82.5 & 	75.1 	& 76.0 \\
  \hline
 \end{tabular}
\footnotemark[1]{\scriptsize{Note that, the COCO trainval set contains \textbf{120k human labeled} images involving 80 object categories. While, our DDT augmentation only depends on \textbf{10k} images of 20 object categories, in especial, these images are \textbf{automatically labeled} by the proposed DDT method.}}
\end{table*}

After that, a state-of-the-art object detection method, i.e., Faster RCNN \citep{fasterrcnn}, is trained as the base model on different training data to validate the effectiveness of DDT augmentation on the object detection task. For the test sets of detection, we employ the \emph{VOC 2007} and \emph{VOC 2012} test set and report the results in Table~\ref{table:webdetect07} and Table~\ref{table:webdetect12}, respectively.

For testing on \emph{VOC 2007}, following \citet{fasterrcnn}, Faster RCNN is trained on ``07+12'' and ``COCO+07+12''. ``07+12'' presents the training data is the union set of \emph{VOC 2007} trainval and \emph{VOC 2012} trainval. ``COCO+07+12'' denotes that except for \emph{VOC 2007} and \emph{VOC 2012}, the \emph{COCO} trainval set is also used for training. ``DDT+07+12'' is our proposal, which uses DDT to process the web images as aforementioned and then combines the processed web data with ``07+12'' as the final training data.

As shown in Table~\ref{table:webdetect07}, our proposal outperforms ``07+12'' by 2.8\% on \emph{VOC 2007}, which is a large margin on the object detection task. In addition, the detection mAP of DDT augmentation is 4\% better than ``07++12'' on the \emph{VOC 2012} test set, cf. Table~\ref{table:webdetect12}. Note that, our DDT augmentation only depends on \textbf{10k} images of 20 object categories, in especial, these images are \textbf{automatically labeled} by the proposed DDT method. 

On the other hand, our mAP is comparable with the mAP training on ``COCO+07+12'' in Table~\ref{table:webdetect07} (or ``COCO+07++12'' in Table~\ref{table:webdetect12}). Here, we would like to point out that the \emph{COCO} trainval set contains \textbf{120k human labeled} images involving 80 object categories, which requires much more human labors, capital and time costs than our DDT augmentation. Therefore, these detection results could validate the effectiveness of DDT augmentation on the object detection task.

\begin{table*}[t!]
 \caption{Comparisons of detection results on the \emph{VOC 2012} test set. Note that, ``07++12'' presents the training data is the union set of VOC 2007 trainval+test and VOC 2012 trainval. ``COCO'' denotes that the COCO trainval set is used for training. ``DDT'' denotes that the webly data processed by DDT augmentation is used for training.} \label{table:webdetect12}
 \centering
 \setlength{\tabcolsep}{0.2pt}
 \begin{tabular}{l||c|c|c|c|c|c|c|c|c|c|c|c|c|c|c|c|c|c|c|c||c}
  \hline
  {Data}  & {aero} & {bike} & {bird}& {boat}& {bottle}& {bus}& {car}& {cat}& {chair}& {cow}& {table}& {dog}& {horse}& {mbike}& {person}& {plant}& {sheep}& {sofa}& {train}& {tv} & mAP (\%) \\  
  \hline
  07++12 & 84.9  & 79.8  & 74.3  & 53.9  & 49.8  & 77.5  & 75.9  & 88.5  & 45.6  & 77.1  & 55.3  & 86.9  & 81.7  & 80.9  & 79.6  & 40.1  & 72.6  & 60.9 &  81.2 &  61.5 & 70.4 \\
  COCO\footnotemark[1]+07++12 & 87.4  & 83.6  & 76.8  & 62.9  & 59.6  & 81.9  & 82.0  & 91.3  & 54.9  & 82.6  & 59.0  & 89.0  & 85.5  & 84.7  & 84.1  & 52.2 &  78.9  & 65.5  & 85.4  & 70.2 & 75.9 \\
  \hline
  DDT+07++12  & 86.5 	 & 81.9 	 & 76.2  & 	63.4  & 	55.4 	 & 80.8  & 	80.1  & 	89.7 	 & 51.6  & 	78.6 	 & 56.2 	 & 88.8 	 & 84.8 	 & 85.5  & 	82.6  & 	50.6  & 	78.1 	 & 64.1 	 & 85.6  & 	68.1  & 	74.4  \\
  \hline
 \end{tabular}
\footnotemark[1]{\scriptsize{Note that, the COCO trainval set contains \textbf{120k human labeled} images involving 80 object categories. While, our DDT augmentation only depends on \textbf{10k} images of 20 object categories, in especial, these images are \textbf{automatically labeled} by the proposed DDT method.}}
\end{table*}

\subsection{Further Study}\label{sec:future}

\begin{figure}[t]
 \centering
 \subfloat[\emph{Dog}]  { \includegraphics[width=\columnwidth]{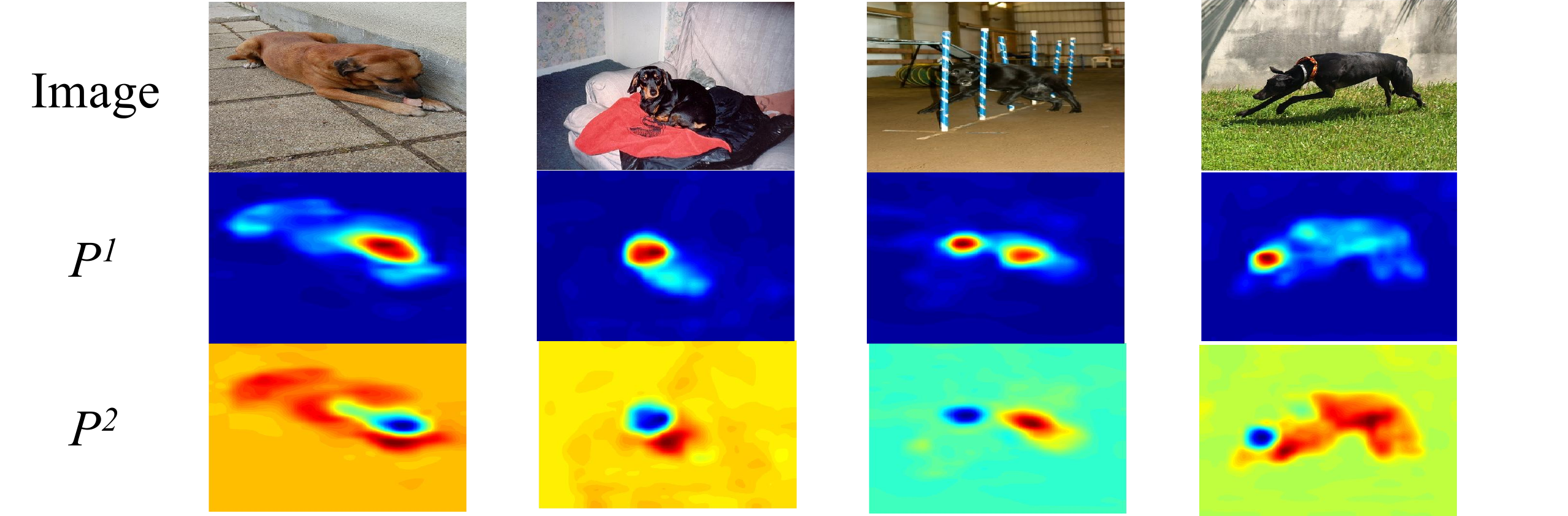} \label{fig:dog1} }

 \subfloat[\emph{Airplane}] { \includegraphics[width=\columnwidth]{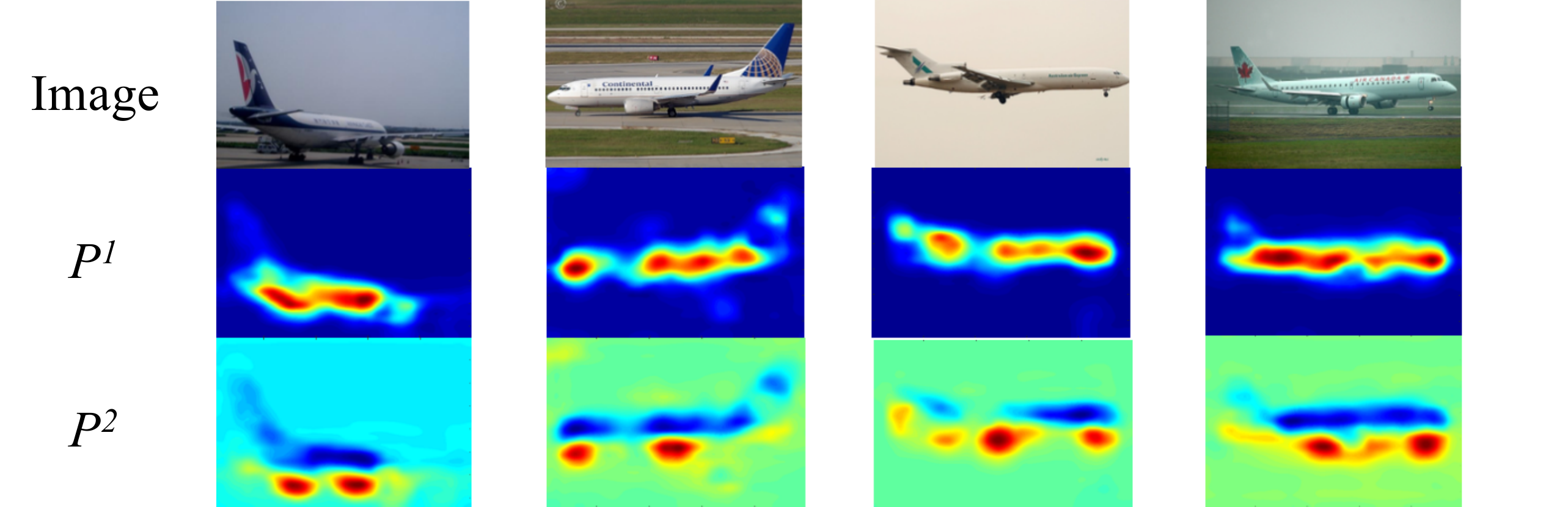} \label{fig:dog2} }

 \subfloat[\emph{Train}] { \includegraphics[width=\columnwidth]{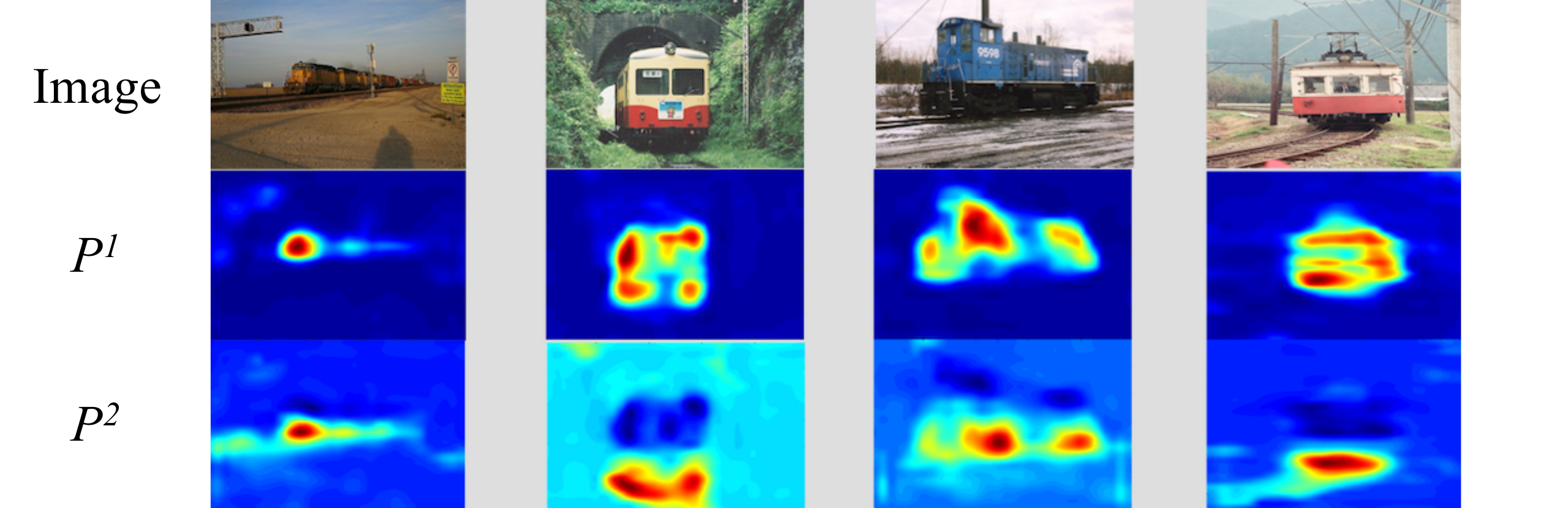} \label{fig:dog3} }
 \caption{Four images belonging to each of three categories of \emph{VOC 2007} with visualization of their indicator matrices $P^1$ and $P^2$. In visualization figures, warm colors indicate positive values, and cool colors present negative. (Best viewed in color.)} \label{fig:dog}
\end{figure}

In the above, DDT only utilizes the information of the first principal components, i.e., $P^1$. How about others, e.g., the second principal components $P^2$? In Fig.~\ref{fig:dog}, we show four images from each of three categories (i.e., dogs, airplanes and trains) in \emph{PASCAL VOC} with the visualization of their $P^1$ and $P^2$. Through these figures, it is apparently to find $P^1$ can locate the whole common object. However, $P^2$ interestingly separates a part region from the main object region, e.g., the head region from the torso region for \textit{dogs}, the wheel and engine regions from the fuselage region for \textit{airplanes}, and the wheel region from the train body region for \textit{trains}. Meanwhile, these two meaningful regions can be easily distinguished from the background. These observations inspire us to use DDT for the more challenging \emph{part-based} image co-localization task in the future, which is never touched before in the literature. 

\section{Conclusions}\label{sec:conc}

Pre-trained models are widely used in diverse applications in computer vision. However, the treasures beneath pre-trained models are not exploited sufficiently. In this paper, we proposed Deep Descriptor Transforming (DDT) for image co-localization. DDT indeed revealed another reusability of deep pre-trained networks, i.e., convolutional activations/descriptors can play a role as a common object detector. It offered further understanding and insights about CNNs. Besides, our proposed DDT method is easy to implement, and it achieved great image co-localization performance. Moreover, the generalization ability and robustness of DDT ensure its effectiveness and powerful reusability in real-world applications. Thus, DDT can be used to handle free but noisy web images and further generate valid data sources for improving both recognition and detection accuracy.

Additionally, DDT also has the potential ability in the applications of video-based unsupervised object discovery. Furthermore, interesting observations in Sec.~\ref{sec:future} make the more challenging but intriguing part-based image co-localization problem be a future work.

\section*{Acknowledgments}
The authors want to thank Yao Li and Bohan Zhuang for conducting part of experiments, and thank Chen-Wei Xie and Hong-Yu Zhou for helpful discussions.

\bibliographystyle{spbasic}      
\bibliography{coloc}

\end{document}